\documentclass{article}

\usepackage{microtype}
\usepackage{graphicx}
\usepackage{subfigure}
\usepackage{booktabs}
\usepackage{multirow}
\usepackage{float}
\usepackage{tcolorbox}
\usepackage{tabularx}

\usepackage{hyperref}

\usepackage[accepted]{icml2024}

\usepackage{amsmath}
\usepackage{amssymb}
\usepackage{mathtools}
\usepackage{amsthm}

\usepackage[capitalize,noabbrev]{cleveref}

\theoremstyle{plain}

\theoremstyle{definition}

\theoremstyle{remark}

\usepackage[textsize=tiny]{todonotes}

\icmltitlerunning{MEQA}

\begin{document}

\twocolumn[
\icmltitle{MEQA: A Meta-Evaluation Framework\\ for Question
\& Answer LLM Benchmarks}
\icmlsetsymbol{equal}{*}

\begin{icmlauthorlist}
\icmlauthor{Jaime Raldua Veuthey}{apart}
\icmlauthor{Zainab Ali Majid}{apart}
\icmlauthor{Suhas Hariharan}{londonuni,apart}
\icmlauthor{Jacob Haimes}{apart}
\end{icmlauthorlist}

\icmlaffiliation{londonuni}{University College London}
\icmlaffiliation{apart}{Apart Research}

\icmlcorrespondingauthor{Jaime Raldua Veuthey}{jaime.raldua.veuthey@gmail.com}

\vskip 0.3in
]

\printAffiliationsAndNotice{}

\begin{abstract}
As Large Language Models (LLMs) advance, their potential for widespread societal impact grows simultaneously. Hence, rigorous LLM evaluations are both a technical necessity and social imperative. While numerous evaluation benchmarks have been developed, there remains a critical gap in meta-evaluation: effectively assessing benchmarks' quality. We propose MEQA, a framework for the meta-evaluation of question and answer (QA) benchmarks, to provide standardized assessments, quantifiable scores, and enable meaningful intra-benchmark comparisons. We demonstrate this approach on cybersecurity benchmarks, using human and LLM evaluators, highlighting the benchmarks' strengths and weaknesses. We motivate our choice of test domain by AI models' dual nature as powerful defensive tools and security threats.
\end{abstract}

\section{Introduction}
Robust LLM evaluations are challenging \cite{biderman2024lessonstrenchesreproducibleevaluation} and past meta-evaluation work has focused on a limited number of specific metrics \cite{biderman2024lessonstrenchesreproducibleevaluation, ren2024safetywashingaisafetybenchmarks}. We propose MEQA, a holistic approach that synthesizes knowledge from previous work \cite{biderman2024lessonstrenchesreproducibleevaluation, ren2024safetywashingaisafetybenchmarks, sclar2024quantifyinglanguagemodelssensitivity, stureborg2024largelanguagemodelsinconsistent, subramonian2023takestangonavigatingconceptualizations} into eight key criteria, described in Table~\ref{criteria-table}. In order to ensure scoring is intuitive and specific, we break down these criteria into 44 sub-criteria, detailed in Appendix~\ref{criteria-sub-criteria}. Evaluators then give scores between 1 and 5 for each sub-criterion.

\begin{table}
  \caption{Meta-evaluation criteria}
  \label{criteria-table}
  \centering
  \begin{tabularx}{.5\textwidth}{@{}>{\raggedright\arraybackslash}p{0.13\textwidth}@{}p{0.01\textwidth}@{}>{\raggedright\arraybackslash}p{0.36\textwidth}@{}}
    \toprule
    \textbf{Criteria} && \textbf{Description} \\
    \midrule
    Memorization Robustness && Tests for abilities of models beyond memorizing information \cite{yu2023skillmixflexibleexpandablefamily, zhu2024dyvaldynamicevaluationlarge}. Publicly available questions, for example, would be insufficient.\vspace{.5em}\\
    Prompt Robustness && Tests for models' consistency in responding to various prompt formats \cite{sclar2024quantifyinglanguagemodelssensitivity}; multi-prompt evaluations \cite{mizrahi2024stateartmultipromptllm}, for example, are effective. \vspace{.5em}\\
    Evaluation Design && Separates evaluation criteria (e.g., technical accuracy, clarity, relevance) and uses appropriate scoring granularity for nuanced assessment \cite{stureborg2024largelanguagemodelsinconsistent, zheng2023judgingllmasajudgemtbenchchatbot, saphra2024tragedyparsehistoryrepeats}. \vspace{.5em}\\
    Evaluator Design && Provides clear evaluation guidelines including considerations on multi-attribute judgment design, annotator disagreement, and ensuring evaluator expertise \cite{fluri2023evaluatingsuperhumanmodelsconsistency, stureborg2024largelanguagemodelsinconsistent}. \vspace{.5em}\\
    Reproducibility && Ensure features such as full transparency and availability of evaluation code. \vspace{.5em}\\
    Comparability && Facilitate consistent and fair comparisons through factors such as standardized task implementations and prompt formats \cite{DBLP:journals/corr/abs-2202-01279}. \vspace{.5em}\\
    Validity && Features such as ensuring representative data and real-world applicability \cite{subramonian2023takestangonavigatingconceptualizations, DBLP:journals/corr/abs-1912-05511}. \vspace{.5em}\\
    Reliability && Demonstrates statistical rigor through output sharing, hyperparameter reporting, multiple runs, confidence intervals, standard errors, and reliability metrics \cite{xiao2023evaluatingevaluationmetricsframework}. \\
    \bottomrule
  \end{tabularx}
\end{table}

\section{Cybersecurity benchmarks}
\label{section-2}

\subsection{Methodology}

We test MEQA using both human and LLM evaluators to score cybersecurity benchmarks. We had three independent human evaluators; their scores agreed exactly on the majority (over 80\%) of the sub-criteria, and disagreements were no more than 1 point apart. 

Additionally, we used GPT-4o to assess benchmarks by prompting with sub-criteria definitions (code is available \href{https://anonymous.4open.science/r/meqa-C46B/README.md}{anonymously} and compute details are present in Appendix~\ref{experimental-details}. The automated scoring tended to match human evaluators, particularly when producing extreme scores (1 or 5). Please see Appendix~\ref{limitations} for full limitations.  This LLM-aided approach undoubtedly saved time and is crucial for scalable, automated meta-evaluations, needed to keep pace with new benchmarks across domains.

\subsection{Results}

\begin{figure*}
  \centering
  \setlength{\abovecaptionskip}{0.1ex}  
  \includegraphics[width=\textwidth]{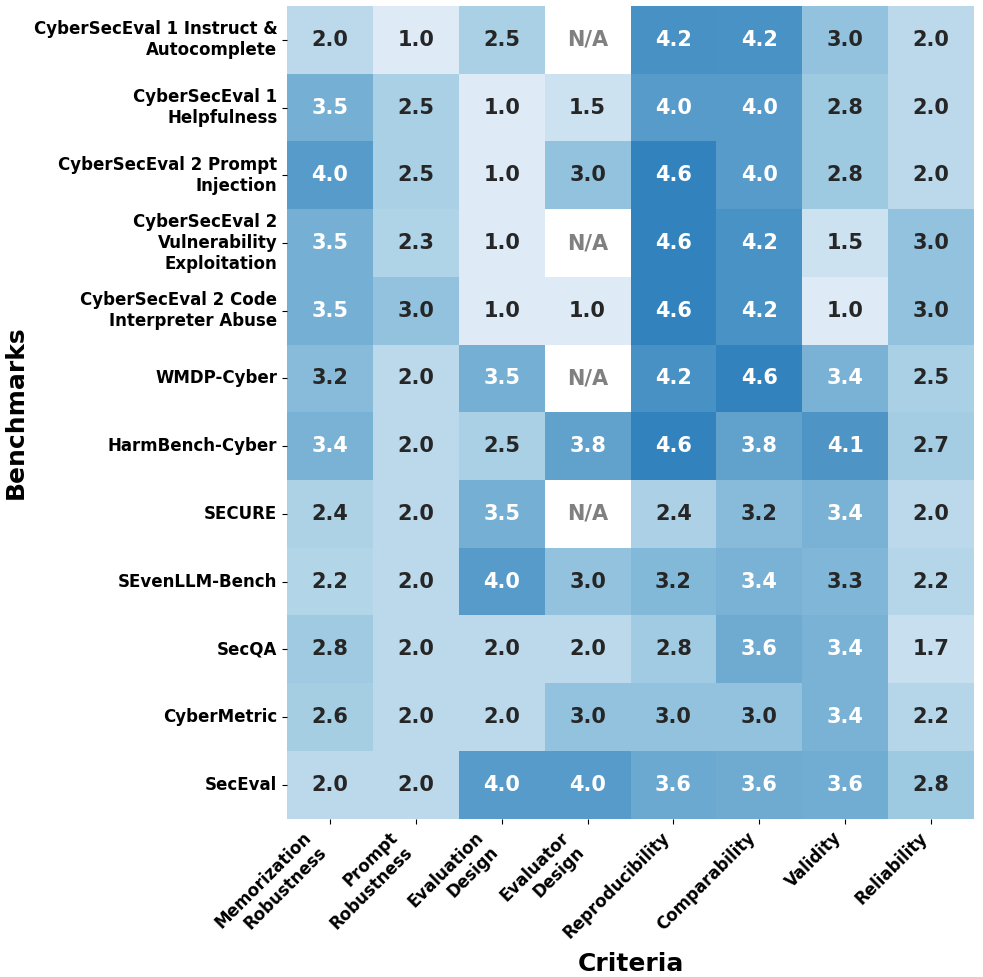}
  \caption{Scores of cybersecurity benchmarks per criterion. N/A indicates inapplicable criteria (e.g. SECURE uses pre-defined correct answers; evaluator design does not apply).}
  \label{fig:scores}
\end{figure*}

Our scores in Figure~\ref{fig:scores} highlight cybersecurity benchmarks’ strengths. Most benchmarks score highly in terms of reproducibility and comparability. SecEval stands out regarding evaluation and evaluator design. HarmBench-Cyber and WMDP-Cyber scored well across several categories and highest scores (Table~\ref{cybersec-eval-results}), indicating relatively high quality.

However, key weaknesses were also brought to light. Most benchmarks obtain low scores for prompt robustness and reliability. The high standard deviations shown in Table~\ref{cybersec-eval-results} reveal significant variability, suggesting that each benchmark's performance fluctuates considerably across different sub-criteria. Scores can be contextualised further using sub-criteria detailed in Appendix~\ref{criteria-sub-criteria} and results in Appendices~\ref{sub-criteria-results} and~\ref{error-analysis}.

\section{Conclusions}

\begin{table}
  \setlength{\belowcaptionskip}{.75\baselineskip}
  \caption{Benchmark scores}
  \label{cybersec-eval-results}
  \centering
  \begin{tabularx}{0.37\textwidth}{@{}l@{}>{\raggedleft\arraybackslash}X@{}}
    \toprule
    \textbf{Benchmark} & \textbf{Mean~(Std)} \\
    \midrule
    HarmBench-Cyber {\small\cite{harmbench}}& 3.6 (1.3) \\
    WMDP-Cyber {\small\cite{wmdp}}& 3.5 (1.6) \\
    CyberSecEval 2 {\small\cite{cse2}}& 3.4 (1.5) \\
    CyberSecEval 1 {\small\cite{cse1}}& 3.2 (1.4) \\
    SecEval {\small\cite{li2023seceval}}& 3.2 (1.4) \\
    SEvenLLM-Bench {\small\cite{seven}}& 2.9 (1.0) \\
    CyberMetric {\small\cite{cybermetric}}& 2.8 (1.2) \\
    SECURE {\small\cite{secure}}& 2.7 (1.1) \\
    SecQA {\small\cite{secqa}}& 2.7 (1.2) \\
    \bottomrule
  \end{tabularx}
\end{table}

We propose a novel meta-evaluation framework, MEQA, and test it on cybersecurity QA benchmarks. While many bechmarks perform well in terms of reproducibility and comparability, there is a clear need for improvement in prompt robustness and reliability. Additionally, our tests using LLM evaluators show promise in improving scalability. We aim to further refine our evaluation criteria and test additional benchmarks. We hope this framework can serve as a blueprint for benchmark developers and enable effective gap-analysis.

\section*{Impact Statement}
\label{sec:impact}
Developing meta-evaluation frameworks for LLM benchmarks requires thoughtful consideration and ongoing refinement. It is crucial to carefully select and regularly update the criteria used. The rigor of these meta-evaluations is of utmost importance, as they effectively establish the standards for both LLM benchmarks and, by extension, the LLMs themselves. By maintaining high-quality meta-evaluation frameworks, we can drive the development of more robust, relevant, and reliable benchmarks, which in turn will lead to the creation of more capable and trustworthy language models. This process of continuous improvement and careful curation of evaluation criteria is essential for advancing the field of AI in a responsible and effective manner.

\section*{Acknowledgements}
We would like to thank Apart Research for supporting our research and hosting the hackathons that initiated this project.

{\small
\bibliography{citations}
\bibliographystyle{icml2024}
}

\newpage
\appendix
\onecolumn
\section*{\centering \Large Appendices}

\section{Criteria and sub-criteria}
\label{criteria-sub-criteria}
We outline our meta-evaluation criteria in Table~\ref{criteria-table}. The below tables add additional detail to this framework, focusing on the 44 sub-criteria and our scoring rubric for each one.

\subsection{Memorization robustness}
The benchmark should be designed so that LLM capabilities being measured go beyond the capacity of LLMs to memorise information. This error mode can happen if the questions in the benchmark are designed using data on which the LLM was originally trained on. Sub-criteria are detailed in Table~\ref{memorization-robustness-table}.

\begin{table}[h]
  \caption{Memorization Robustness sub-criteria}
  \label{memorization-robustness-table}
  \centering
  \small
  \begin{tabularx}{\textwidth}{@{}>{\raggedright\arraybackslash}p{0.18\textwidth}>{\raggedright\arraybackslash}X@{}}
    \toprule
    \textbf{Sub-criteria} & \textbf{Detail} \\
    \midrule
    Training cut-off date & 
    \textbf{Purpose:} The benchmark should include questions that are unlikely to be in the model's training data. \\
    & \textbf{Low score:} Uses questions from popular certification tests (likely in LLM training data). \\
    & \textbf{High score:} Uses new information sources published after the LLM training cut-off date; questions are curated by domain experts based on this new information. \\
    \addlinespace[1ex]
    Public/private data & 
    \textbf{Purpose:} Check if questions are taken from open public data or from private sources. \\
    & \textbf{Low score:} Questions are from very easily accessible public data (e.g. Wikipedia). \\
    & \textbf{Medium score:} Questions are from less accessible public sources (e.g. behind paywall). \\
    & \textbf{High score:} Questions are from private, proprietary data or non-digitalized data. \\
    \addlinespace[1ex]
    Question Formulation & 
    \textbf{Purpose:} Look at how questions are formulated: copied directly from the original source, paraphrased, or newly curated by domain experts. \\
    & \textbf{Low score:} Questions are directly copied from certification test exams. \\
    & \textbf{High score:} Questions are modified (e.g. altered from multiple-choice to open-ended, or from direct recall to scenario-based problems). \\
    \addlinespace[1ex]
    Dynamic Generation & 
    \textbf{Purpose:} The benchmark uses dynamic generation techniques to create evaluation data. These methods create scenarios extremely unlikely to have been seen in training data, ensuring tests of true understanding and skill combination rather than mere memorization. \\
    & \textbf{Low score:} Uses static dataset of pre-defined questions. \\
    & \textbf{High score:} Employs dynamic generation methods such as combinatorial skill mixing (e.g. \textit{Skill-Mix} approach \cite{yu2023skillmixflexibleexpandablefamily}) or graph-informed dynamic evaluation (e.g. \textit{DyVal} approach \cite{zhu2024dyvaldynamicevaluationlarge}). \\
    \addlinespace[1ex]
    Memorization Detection & 
    \textbf{Purpose:} Check if the benchmark implements specific techniques to detect and quantify memorization. \\
    & \textbf{Low score:} No techniques implemented to detect or measure potential memorization. \\
    & \textbf{High score:} Implements techniques like textit{``canary strings''} to prove absence of memorization issues, and provides quantifiable measures of any potential memorization. \\
    \bottomrule
  \end{tabularx}
\end{table}
\vfill
\pagebreak

\subsection{Prompt robustness}
A measure of prompt sensitivity and consistency, sub-criteria detailed in Table~\ref{prompt-robustness-criteria-table}. Prompt robustness aims to measure if benchmarks test LLMs' abilities to understand the intent behind a prompt, and provide reliable responses.

\begin{table}[h]
  \caption{Prompt Robustness sub-criteria}
  \label{prompt-robustness-criteria-table}
  \centering
  \small
  \begin{tabularx}{\textwidth}{@{}>{\raggedright\arraybackslash}p{0.18\textwidth}>{\raggedright\arraybackslash}X@{}}
    \toprule
    \textbf{Sub-criteria} & \textbf{Detail} \\
    \midrule
    Prompt Diversity & 
    \textbf{Purpose:} Testing models on format variations of questions to see if they reply consistently. \\
    & \textbf{Low score:} Benchmark with only one prompt per task; each question appears once without duplicated paraphrased questions. \\
    & \textbf{Medium score:} Benchmark uses a systematic approach like \textit{FormatSpread} to evaluate performance across a range of plausible prompt formats (efficiently sampled from the prompt design space). \\
    & \textbf{High score:} Benchmark with 50+ diverse prompts per task, including rephrases, chain-of-thought, and gradual prompts (uses multi-prompt evaluation). \\
    \addlinespace[1ex]
    Analysis of Prompt Variations & 
    \textbf{Purpose:} Assessment of how responses vary per prompt as in the \textit{Multi-prompt LLM evaluation} approach\cite{mizrahi2024stateartmultipromptllm}. Note that this only applies if there is prompt diversity. \\
    & \textbf{Low score:} No analysis of performance variation across prompts. \\
    & \textbf{High score:} Comprehensive analysis of model ranking changes and performance variability across prompts. \\
    \addlinespace[1ex]
    Multi-prompt Evaluation Metrics & 
    \textbf{Purpose:} Include metrics based on multiple prompts such as those detailed in the \textit{Multi-Prompt LLM evaluation} approach\cite{mizrahi2024stateartmultipromptllm}. \\
    & \textbf{Low score:} Only reporting accuracy on the best-performing prompt. \\
    & \textbf{High score:} Reporting MaxP, AvgP, Sat, and CPS metrics for each model and task. Report distributions of scores across different prompting setups. \\
    \addlinespace[1ex]
    Prompt Quality & 
    \textbf{Purpose:} Assess the quality of prompts, for example, in terms of clarity, language, and grammar. \\
    & \textbf{Low score:} Prompts with grammatical errors or unclear instructions. \\
    & \textbf{High score:} Manually verified prompts adhering to strict PromptSource quality guidelines. \\
    \bottomrule
  \end{tabularx}
\end{table}
\vfill
\subsection{Evaluation Design}
This focuses on assessing the design of the evaluations. Benchmarks should not collapse scores for different features, and scoring should be granular enough to allow for nuanced assessments, as detailed in Table~\ref{evaluation-methodology-criteria-table}.

\begin{table}[h]
  \caption{Evaluation Design sub-criteria}
  \label{evaluation-methodology-criteria-table}
  \centering
  \small
  \begin{tabularx}{\textwidth}{@{}>{\raggedright\arraybackslash}p{0.18\textwidth}>{\raggedright\arraybackslash}X@{}}
    \toprule
    \textbf{Sub-criteria} & \textbf{Detail} \\
    \midrule
    Evaluation Criteria Separation & 
    \textbf{Purpose:} The benchmark should separate different aspects of evaluation to provide a more nuanced and accurate assessment, and ensure distinct features are not collapsed into a single metric\cite{saphra2024tragedyparsehistoryrepeats, clark-etal-2021-thats}.\\
    & \textbf{Low score:} A benchmark that combines technical accuracy, writing style, and practical applicability into a single score, obscuring which features specifically are strong or weak. \\
    & \textbf{High score:} A benchmark that provides separate scores for different criteria such as technical correctness, clarity of explanation, relevance to the question, and practical applicability in real-world contexts. \\
    \addlinespace[1ex]
    Scoring Granularity & 
    \textbf{Purpose:} The benchmark should use an appropriate scoring scale that allows for meaningful differentiation between responses. Optimal granularity may depend on the specific task and context however, a 1-10 scale has worked well previously. \cite{stureborg2024largelanguagemodelsinconsistent}. \\
    & \textbf{Low score:} A benchmark that uses a binary (correct/incorrect) or very limited (e.g. 1-3) scoring scales, that do not allow for nuanced evaluations. Or, a benchmark that uses too broad a scale (e.g. 1-100) that LLMs struggle to use consistently. \\
    & \textbf{High score:} A benchmark that uses a 1-10 scoring scale, providing enough granularity to differentiate between varying levels of response quality, while avoiding the pitfalls of overly broad scales such as round number bias \cite{stureborg2024largelanguagemodelsinconsistent}. \\
    \bottomrule
  \end{tabularx}
\end{table}
\pagebreak

\subsection{Evaluator Design}
This criterion is only relevant to benchmarks that require either human or LLM evaluators to give scores. There are other ways to implement benchmarks that do not require this; for example, using heuristics or logic to evaluate answers. We assess evaluator design through the sub-criteria detailed in Table~\ref{evaluation-methodology-extended-criteria-table}.

\begin{table}[h]
  \caption{Evaluator design sub-criteria}
  \label{evaluation-methodology-extended-criteria-table}
  \centering
  \small
  \begin{tabularx}{\textwidth}{@{}>{\raggedright\arraybackslash}p{0.18\textwidth}>{\raggedright\arraybackslash}X@{}}
    \toprule
    \textbf{Sub-criteria} & \textbf{Detail} \\
    \midrule
    Evaluation Guidelines & 
    \textbf{Purpose:} Benchmarks should be carefully designed to provide meaningful and consistent results. \\
    & \textbf{Low score:} A benchmark that asks human evaluators to rate responses on a single``good to bad'' scale without any specific criteria or guidelines, leading to inconsistent and subjective evaluations. \\
    & \textbf{High score:} A benchmark that provides clear, detailed rubrics for human evaluators, breaking down the assessment into specific aspects such as technical accuracy, clarity of explanation, and practical applicability, with concrete examples for each rating level. \\
    \addlinespace[1ex]
    Multi-Attribute Judgment Design & 
    \textbf{Purpose:} If the benchmark evaluates multiple attributes of a response, it should account for potential anchoring effects, where earlier scores tend to influence subsequent scores. Please note that this effect is more likely for LLM evaluators, rather that human evaluators\cite{stureborg2024largelanguagemodelsinconsistent}. \\
    & \textbf{Low score:} A benchmark that asks the evaluator to judge multiple attributes (e.g., accuracy, relevance, and clarity) in a single generation, without considering how earlier judgments might influence later ones. \\
    & \textbf{High score:} A benchmark that either evaluates each attribute in separate generations or implements a randomized order of attribute evaluation, along with statistical controls to detect and mitigate anchoring effects between judgments. \\
    \addlinespace[1ex]
    Handling Annotator Disagreement & 
    \textbf{Purpose:} If there is more than one evaluator, the benchmark should have a clear strategy for addressing and interpreting disagreements between evaluators. \\
    & \textbf{Low score:} A benchmark that simply averages scores from multiple evaluators without considering the reasons for disagreement or the potential validity of minority opinions. \\
    & \textbf{High score:} A benchmark that uses a deliberation process for cases of significant disagreement, allowing evaluators to discuss their reasoning and potentially revise their assessments, or that preserves and reports the distribution of evaluations to reflect genuine differences in expert opinions on complex topics. \\
    \addlinespace[1ex]
    Annotator Expertise & 
    \textbf{Purpose:} The benchmark should ensure that evaluators have the necessary expertise to accurately assess the content. Prefereable evaluators would be more sophisticated LLMs or human domain experts. \\
    & \textbf{Low score:} A benchmark that uses general, crowd-sourcing platforms to evaluate highly technical concepts, resulting in evaluations from individuals who may not understand the subject matter. \\
    & \textbf{High score:} A benchmark that employs certified domain professionals or academics with relevant expertise to evaluate responses, ensuring that the assessments are based on deep domain knowledge. \\
    \addlinespace[1ex]
    Consistency Checking & 
    \textbf{Purpose:} The evaluation process should include methods to check the consistency of the evaluator's decisions across multiple evaluations of the same or similar items. For example, by using redundant questions and answers to check for the consistency of the evaluators' responses. \\
    & \textbf{Low score:} A benchmark that does not include any consistency checks, potentially allowing for unreliable or contradictory judgments. \\
    & \textbf{High score:} A benchmark that regularly includes duplicate or highly similar items in the evaluation set and compares the evaluators' decisions to ensure consistency, flagging and investigating any discrepancies. \\
    \addlinespace[1ex]
    Calibration with Human Judgment & 
    \textbf{Purpose:} LLM evaluators' scores should be calibrated against human experts' scores to ensure alignment with human standards. \\
    & \textbf{Low score:} A benchmark that relies solely on LLM scores without any comparison to human expert evaluations. \\
    & \textbf{High score:} A benchmark that regularly compares LLM scores with those of domain experts, calculating agreement rates and adjusting the LLM evaluation process to better align with human judgments.\\
    \bottomrule
  \end{tabularx}
\end{table}
\vfill
\pagebreak

\subsection{Reproducibility}
Ensure benchmarks can be reproduced easily: code, prompts and datasets should be readily available. Instructions and evaluation metrics should be described in detail. A complete list of sub-criteria is present in Table~\ref{code-evaluation-criteria-table}.

\begin{table}[h]
  \caption{Reproducibility sub-criteria}
  \label{code-evaluation-criteria-table}
  \centering
  \small
  \begin{tabularx}{\textwidth}{@{}>{\raggedright\arraybackslash}p{0.18\textwidth}>{\raggedright\arraybackslash}X@{}}
    \toprule
    Sub-criteria & Detail \\
    \midrule
    Availability of Evaluation Code & 
    \textbf{Purpose:} Check if code is available to ensure the benchmark can be reproduced easily. \\
    & \textbf{Low score:} No evaluation code provided, only high-level descriptions in the paper. \\
    & \textbf{High score:} Full evaluation code publicly available on GitHub with clear documentation. \\
    \addlinespace[1ex]
    Versioning & 
    \textbf{Purpose:} Assess versioning (using GitHub history). Please note that this only applies if the benchmark is publicly available and maintained. \\
    & \textbf{Low score:} No version control, frequent undocumented changes to the benchmark. \\
    & \textbf{High score:} Clear versioning system (e.g. semantic versioning) with a change log for each update. \\
    \addlinespace[1ex]
    Prompt Transparency & 
    \textbf{Purpose:} Ensure adequate prompt transparency to allow for experiments to be repeated. \\
    & \textbf{Low score:} Prompts are not disclosed or only vaguely described. \\
    & \textbf{High score:} All prompts used in the benchmark are explicitly provided. \\
    \addlinespace[1ex]
    Data Accessibility & 
    \textbf{Purpose:} Required datasets should be accessible for the benchmark to be reproduced. \\
    & \textbf{Low score:} Proprietary or inaccessible datasets used for evaluation. \\
    & \textbf{High score:} All datasets used are publicly available or clearly described for recreation. \\
    \addlinespace[1ex]
    Reproducibility Instructions & 
    \textbf{Purpose:} Instructions to reproduce the evaluations should be available and intelligible. \\
    & \textbf{Low score:} No instructions on how to reproduce results. \\
    & \textbf{High score:} Step-by-step guide for reproducing results, including environment setup and potential pitfalls. \\
    \addlinespace[1ex]
    Evaluation Metric Transparency & 
    \textbf{Purpose:} Any evaluation metrics used should be explained in detail, to allow for a high level of understanding. \\
    & \textbf{Low score:} Vague descriptions of how metrics are calculated. \\
    & \textbf{High score:} Detailed explanations of all evaluation metrics, including edge case handling. \\
    \addlinespace[1ex]
    Baseline Model Inclusion & 
    \textbf{Purpose:} Baseline models should be included as part of the benchmark methodology, with the ability for reproduction. \\
    & \textbf{Low score:} No baseline models provided for comparison. \\
    & \textbf{High score:} Several well-documented baseline models included, with instructions for running them. \\
    \addlinespace[1ex]
    Scalability and Efficiency & 
    \textbf{Purpose:} Ability to reproduce this work without particularly time or resource intensive processes. \\
    & \textbf{Low score:} It takes several months of time and significant GPU resources to benchmark a single model. \\
    & \textbf{High score:} Can evaluate multiple models on thousands of prompts within hours. \\
    \bottomrule
  \end{tabularx}
\end{table}
\vfill
\pagebreak

\subsection{Comparability}
Ensure comparisons are consistent and fair, and benchmarks are not biased or limited to specific models or architectures. Specific details are in Table~\ref{implementation-criteria-table}.

\begin{table}[h]
  \caption{Comparability sub-criteria}
  \label{implementation-criteria-table}
  \centering
  \small
  \begin{tabularx}{\textwidth}{@{}>{\raggedright\arraybackslash}p{0.18\textwidth}>{\raggedright\arraybackslash}X@{}}
    \toprule
    Sub-criteria & Detail \\
    \midrule
    Standardized Task Implementations & 
    \textbf{Purpose:} Benchmarks should use standard frameworks such as lm-eval harness \cite{eval-harness} to ensure consistent implementation across different models. \\
    & \textbf{Low score:} Each task has its own unique setup process, making it difficult to run the full benchmark consistently or compare results across models. \\
    & \textbf{High score:} Use standardized implementations of tasks as provided by lm-eval. Use consistent prompt formats and scoring methods across different models. \\
    \addlinespace[1ex]
    Prompt Templating and Formatting & 
    \textbf{Purpose:} If it is not possible to standardise task implementations, benchmarks should at least ensure consistent templating and formatting across questions and tasks. Consistent prompting has previously been demonstrated by the \textit{PromptSource} approach \cite{DBLP:journals/corr/abs-2202-01279}. \\
    & \textbf{Low score:} Inconsistent prompt formats across tasks. \\
    & \textbf{High score:} Standardized templating, for example, Jinja2 \cite{Jinja2}, for all prompts, allowing easy programmatic use. \\
    \addlinespace[1ex]
    Cross-Model Consistency & 
    \textbf{Purpose:} Benchmarks should use the same evaluation setup - prompts, scoring methods, few or zero-shot appraches - for all models being compared. \\
    & \textbf{Low score:} Tasks designed to favor specific model architectures or training paradigms. \\
    & \textbf{High score:} Tasks formulated to be architecture-neutral, testing general language understanding capabilities. \\
    \addlinespace[1ex]
    Adaptability to Different Models & 
    \textbf{Purpose:} Ensure the benchmark is model-agnostic, to allow fair comparisons between a variety of LLMs.\\
    & \textbf{Low score:} The benchmark is only applicable to one specific model. \\
    & \textbf{High score:} The benchmark can be used to evaluate any text-in/text-out language model. \\
    \addlinespace[1ex]
    Cross-Platform Compatibility & 
    \textbf{Purpose:} Benchmarks should not be limited to specific hardware, software or platforms. \\
    & \textbf{Low score:} Benchmark can only run on a specific hardware/software configuration. For example, testing code that can only run on Unix machines. \\
    & \textbf{High score:} Benchmark is designed to be platform-agnostic with clear instructions for different environments. \\
    \bottomrule
  \end{tabularx}
\end{table}
\vfill
\pagebreak

\subsection{Validity}
Ensure that the benchmark assesses what it claims to assess, and provides meaningful results. We include sub-criteria such as face validity, substantive validity, discriminant validity and convergent validity following the approach of \textit{Subramonian et. al} \cite{subramonian2023takestangonavigatingconceptualizations}. Details in Table~\ref{valid-criteria-table}.

\begin{table}[H]
  \caption{Validity sub-criteria}
  \label{valid-criteria-table}
  \centering
  \small
  \begin{tabularx}{\textwidth}{@{}>{\raggedright\arraybackslash}p{0.18\textwidth}>{\raggedright\arraybackslash}X@{}}
    \toprule
    Sub-criteria & Detail \\
    \midrule
    Representative data & 
    \textbf{Purpose:} The data is representative of what the benchmark aims to evaluate. For example, for cybersecurity benchmarks, datasets should be representative of the most common attacks. \\
    & \textbf{Low score:} A benchmark that uses only one book as a source for all its questions and claims to have wide coverage. \\
    & \textbf{High score:} A benchmark focused on vulnerabilty detections specifically, that uses information on existing vulnerabilities. \\
    \addlinespace[1ex]
    Proper elicitation & 
    \textbf{Purpose:} Properly elicit the model before assessing its performance on the benchmark. Evaluate performance across various prompt formats (e.g., zero-shot, few-shot, different instructions). \\
    & \textbf{Low score:} The benchmark tests only for zero-shot and authors give conclusions on upper bounds of capabilities. \\
    & \textbf{Medium score:} The benchmark tests only for zero-shot but authors make it very clear that this has much room for improvement with better elicitation techniques. \\
    & \textbf{High score:} Rigorous use of many different elicitation techniques. \\
    \addlinespace[1ex]
    Metric consistency & 
    \textbf{Purpose:} Compare different metrics (e.g. Rouge-L \cite{lin-2004-rouge}, Semantic Similarity, GPT-4 evaluation, and multiple-choice questions) and ensure their results are consistent. \\
    & \textbf{Low score:} Only uses another LLM to create evaluation scores. \\
    & \textbf{High score:} Uses different metrics and shows that they are consistent or explains the inconsistencies if they exist. \\
    \addlinespace[1ex]
    Real World Task Design & 
    \textbf{Purpose:} The benchmark questions should be systematically chosen and well-defined, reflecting the actual problem space they aim to represent. \\
    & \textbf{Low score:} A benchmark with arbitrarily selected tasks that do not reflect real-world challenges or are based on outdated vulnerabilities. \\
    & \textbf{High score:} A benchmark with tasks systematically designed to cover a range of current and emerging threats, vetted by domain experts and regularly updated. \\
    \addlinespace[1ex]
    Face Validity & 
    \textbf{Purpose:} A preliminary ``sniff test'' that assesses a benchmark on surface level. \\
    & \textbf{Low score:} A benchmark that claims to assess advanced domain knowledge but only includes basic, common-sense questions about computer usage. \\
    & \textbf{High score:} A benchmark with questions that clearly relate to specific, relevant domain concepts and scenarios, reviewed and approved by domain experts.\\
    \addlinespace[1ex]
    Substantive Validity (Coverage) & 
    \textbf{Purpose:} The benchmark should comprehensively cover the full range of skills and knowledge it aims to assess. \\
    & \textbf{Low score:} A benchmark that only focuses on one area. For example, a cybersecurity Q\&A benchmark that only focuses on password security. \\
    & \textbf{High score:} A benchmark that includes a wide range of question types and difficulty levels, covering all major required areas, and regularly updated. \\
    \addlinespace[1ex]
    Discriminant Validity & 
    \textbf{Purpose:} The benchmark should not inadvertently measure unrelated skills or allow for success through unintended methods. \\
    & \textbf{Low score:} A benchmark where high scores can be achieved by simply having better programming skills, rather than demonstrating true domain knowledge. \\
    & \textbf{High score:} A benchmark with carefully crafted questions that require genuine understanding and problem-solving skills, with no shortcuts or loopholes that could be exploited. \\
    \addlinespace[1ex]
    Convergent Validity & 
    \textbf{Purpose:} The benchmark results should align with other accepted measures of field knowledge and skills. \\
    & \textbf{Low score:} A benchmark for which top performers consistently fail real-world field-specific assessments or certifications. \\
    & \textbf{High score:} A benchmark for which performance strongly correlates with success in practical tasks and aligns with assessments by human experts in the field. Multiple human experts agree on the questions. \\
    \bottomrule
  \end{tabularx}
\end{table}
\vfill
\pagebreak

\subsection{Reliability}
Demonstration of statistical rigor, consistency and stability of measurements. Please note the sub-criteria Model Output Sharing and Hyperparameter Reporting are also relevant for Reproducibility. We also include key points following the approach of \textit{Xiao et. al} \cite{xiao2023evaluatingevaluationmetricsframework}: test-retest reliability, internal consistency, and inter-rater reliability. Further details are present in Table~\ref{reliability-table}.

\begin{table}[h]
  \caption{Reliability sub-criteria}
  \label{reliability-table}
  \centering
  \small
  \begin{tabularx}{\textwidth}{@{}>{\raggedright\arraybackslash}p{0.18\textwidth}>{\raggedright\arraybackslash}X@{}}
    \toprule
    Sub-criteria & Detail \\
    \midrule
    Model Output Sharing & 
    \textbf{Purpose:} Ensure model outputs, in addition to final scores are reported, to show consistency.\\
    & \textbf{Low score:} Only final scores reported, no access to model outputs. \\
    & \textbf{High score:} Full model outputs for all evaluated models made publicly available. \\
    \addlinespace[1ex]
    Hyperparameter Reporting & 
    \textbf{Purpose:} Ensure hyperparameters are amongst features reported, to allow for both reliability and reproducibility.\\
    & \textbf{Low score:} No mention of crucial hyperparameters used during evaluation. \\
    & \textbf{High score:} Comprehensive list of all relevant hyperparameters, including sampling settings for generative tasks. \\
    \addlinespace[1ex]
    Statistical Rigor & 
    \textbf{Purpose:} Ensure the benchmark results are statistically sound and reliable. Examples of good practice include: multiple runs with different random seeds; reporting confidence intervals; standard error reporting; sample size transparency; significance testing; effect size reporting; bootstrapping for uncertainty estimation; subgroup analysis; power analysis, and handling of outliers and errors. \\
    & \textbf{Low score:} No reporting of uncertainty, statistical significance or any other statistical metrics. \\
    & \textbf{High score:} Clear reporting of confidence intervals, standard errors, and statistical significance tests where appropriate. \\
    \addlinespace[1ex]
    Test-Retest Reliability & 
    \textbf{Purpose:} Ensuring the consistency of a metric when applied to the same outputs at different times. Evaluated through computing the metric score twice for each model's output and calculating the correlation between the two sets of scores. \\
    & \textbf{Low score:} Correlation coefficient < 0.7, indicating poor stability. \\
    & \textbf{High score:} Correlation coefficient > 0.9, indicating excellent stability. \\
    \addlinespace[1ex]
    Internal Consistency & 
    \textbf{Purpose:} The degree to which different parts of the benchmark consistently measure the same construct, using Cronbach's alpha or other internal consistency measures. \\
    & \textbf{Low score:} Cronbach's alpha < 0.6, indicating poor internal consistency. \\
    & \textbf{High score:} Cronbach's alpha > 0.8, indicating good to excellent internal consistency. \\
    \addlinespace[1ex]
    Inter-Rater Reliability & 
    \textbf{Purpose:} The degree of agreement among different evaluators. Use measures such as Cohen's kappa (for two raters) or Fleiss' kappa (for multiple raters) for categorical judgments, or Intraclass Correlation Coefficient (ICC) for continuous ratings. \\
    & \textbf{Low score:} Kappa or ICC < 0.4, indicating poor agreement. \\
    & \textbf{High score:} Kappa or ICC > 0.75, indicating excellent agreement. \\
    \bottomrule
  \end{tabularx}
\end{table}
\vfill
\pagebreak

\section{Additional results}
\label{sub-criteria-results}

\begin{figure}[h]
    \centering
    \includegraphics[width=.9\textwidth]{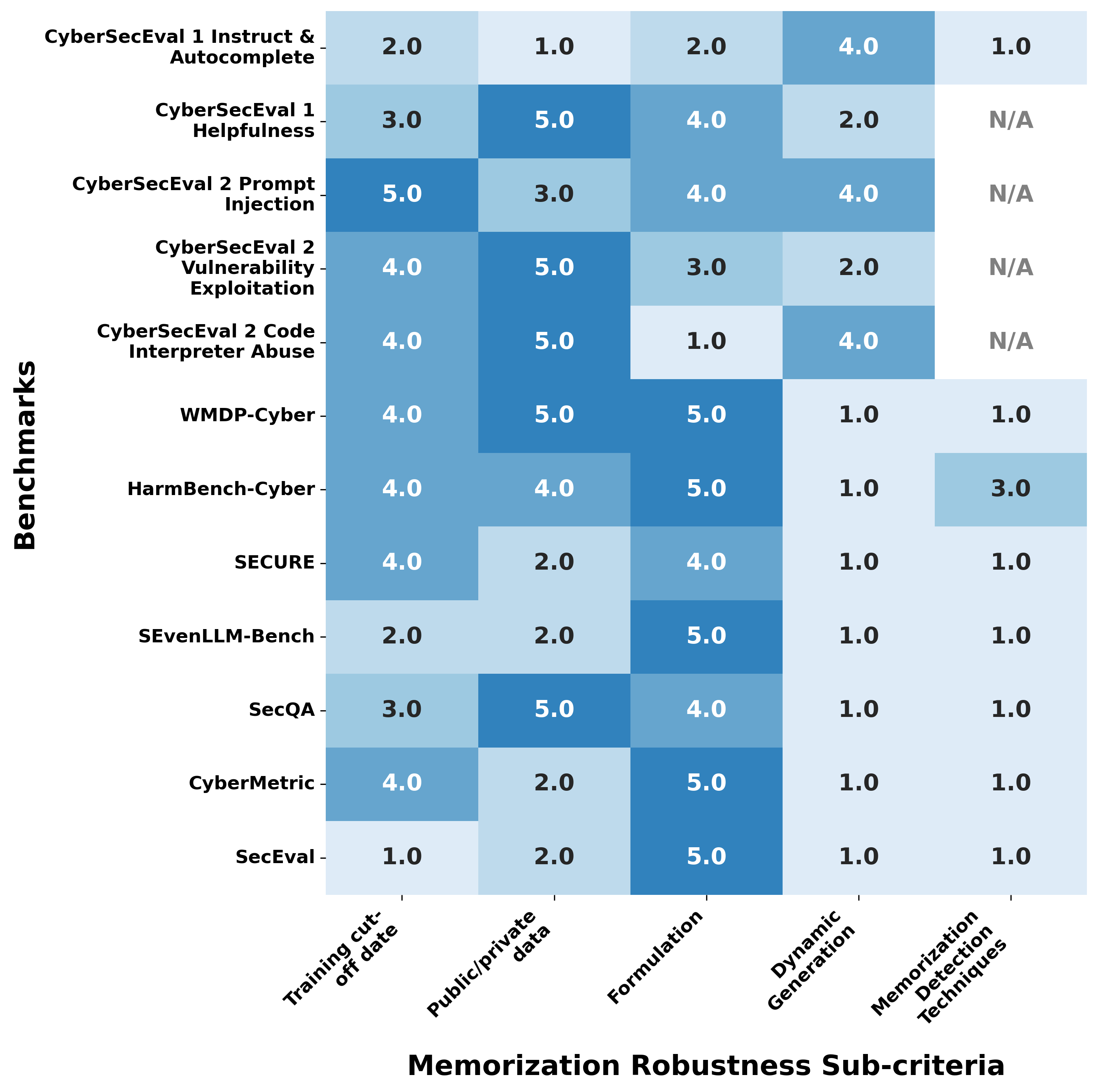}
    \caption{Scores of cybersecurity benchmarks across memorization robustness sub-criteria.}
    \label{fig:memorization}
\end{figure}

\begin{figure}[htbp]
    \centering
    \includegraphics[width=0.9\textwidth]{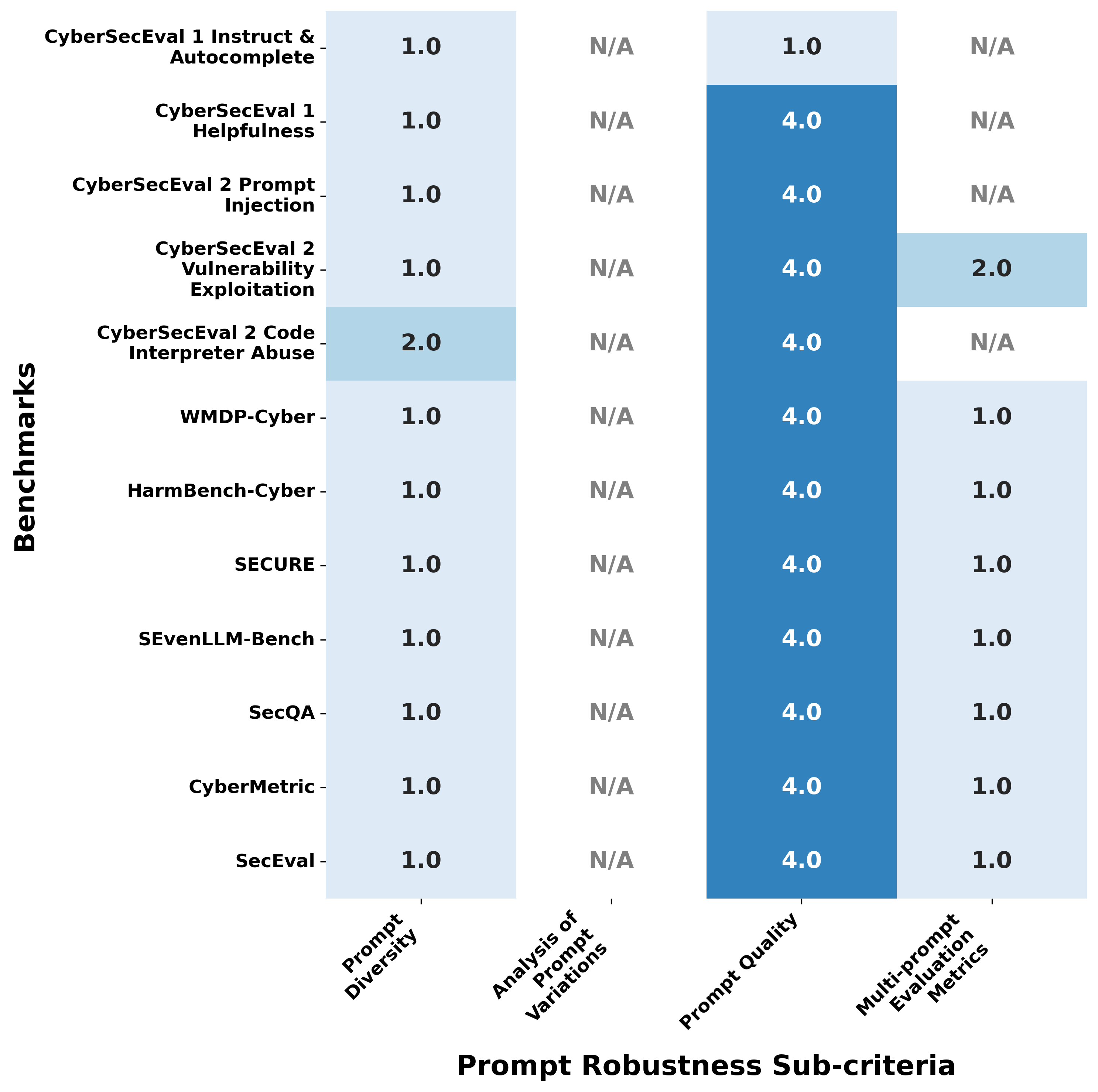}
    \caption{Scores of cybersecurity benchmarks across prompt robustness sub-criteria.}
    \label{fig:promptrobustness}
\end{figure}

\begin{figure}[htbp]
    \centering
    \includegraphics[width=0.9\textwidth]{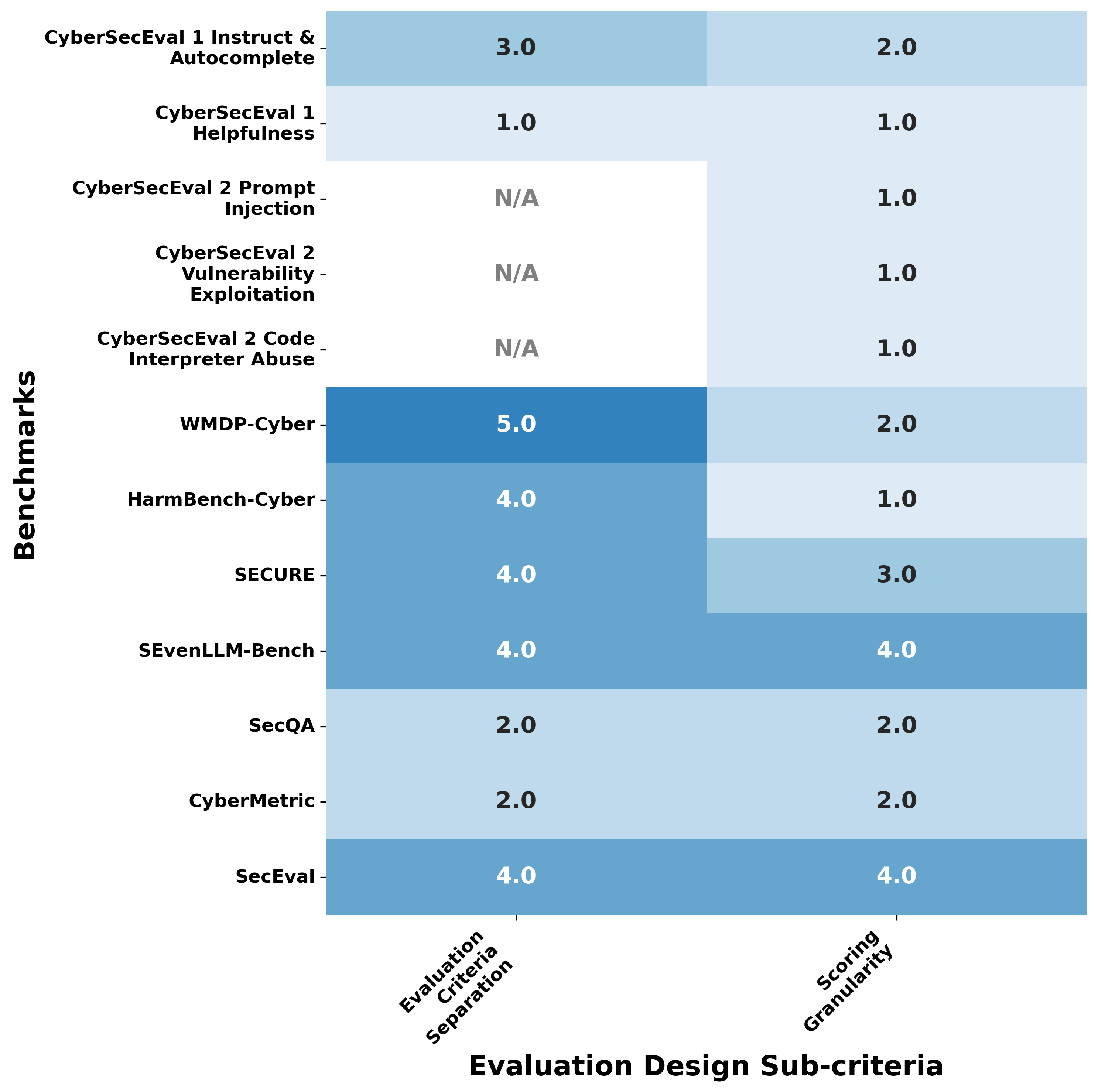}
    \caption{Scores of cybersecurity benchmarks across evaluation design sub-criteria.}
    \label{fig:evaluationdesign}
\end{figure}

\begin{figure}[htbp]
    \centering
    \includegraphics[width=0.9\textwidth]{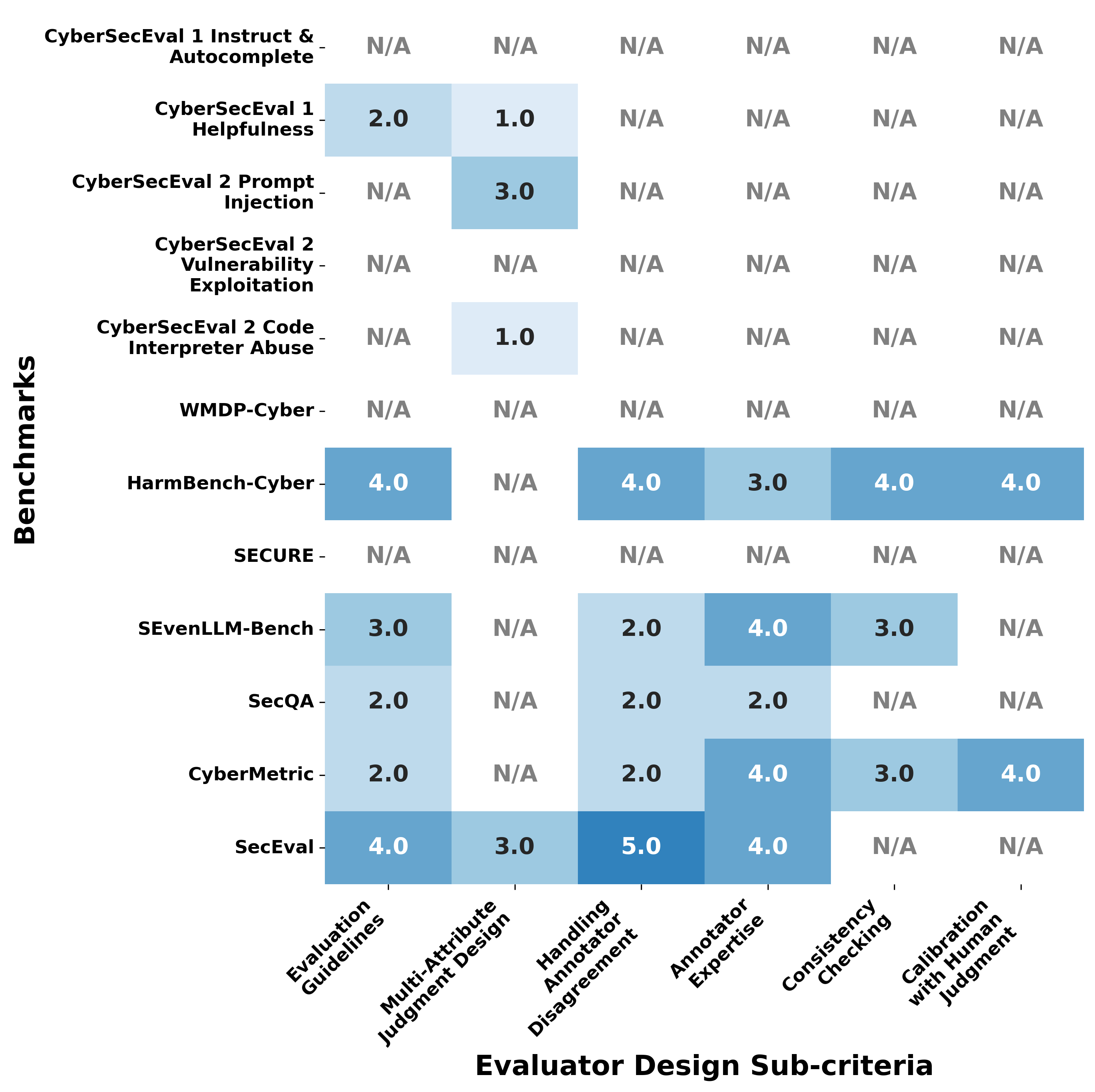}
    \caption{Scores of cybersecurity benchmarks across evaluator design sub-criteria.}
    \label{fig:evaluatordesign}
\end{figure}

\begin{figure}[htbp]
    \centering
    \includegraphics[width=0.9\textwidth]{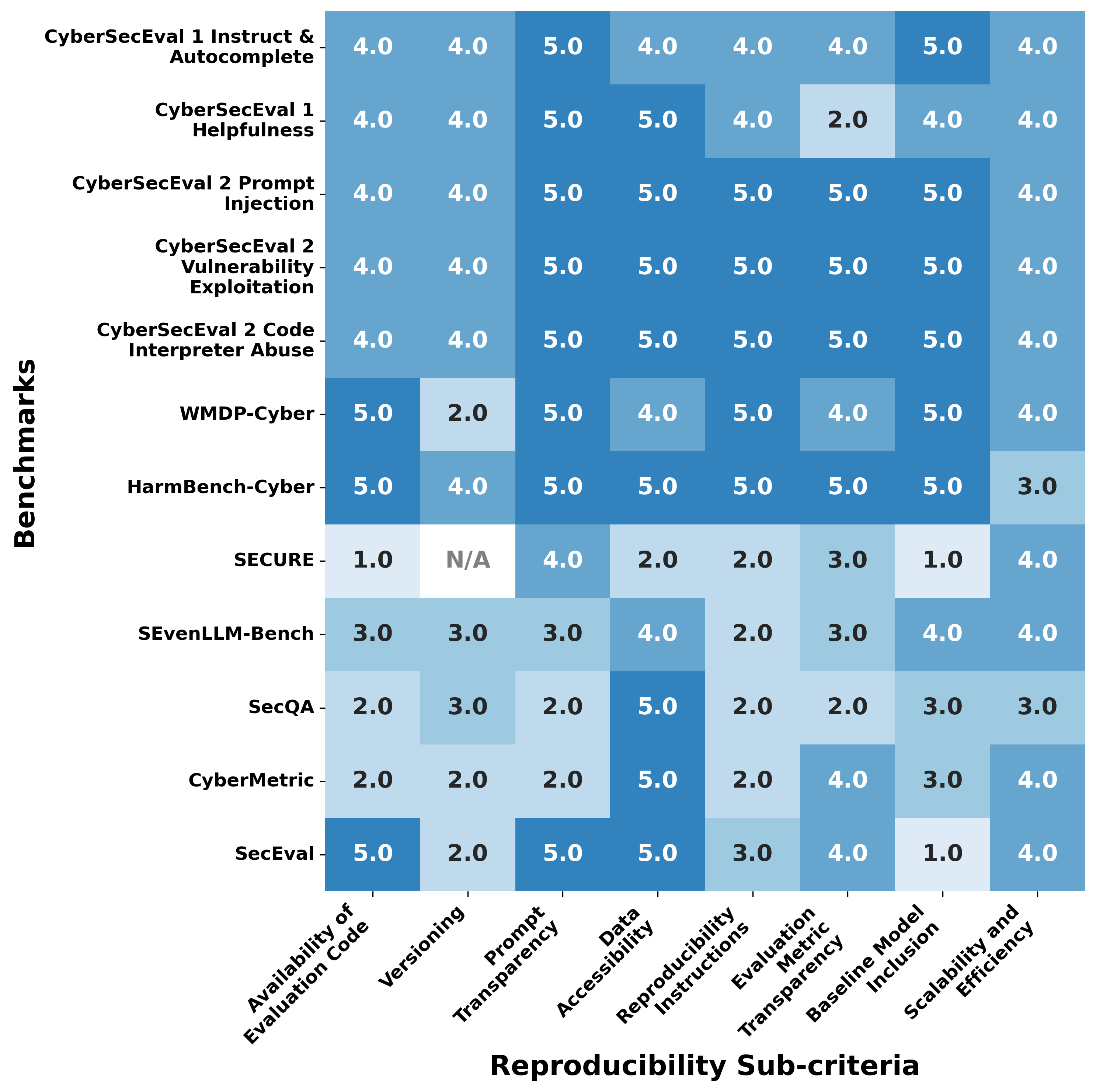}
    \caption{Scores of cybersecurity benchmarks across reproducibility sub-criteria.}
    \label{fig:reproducibility}
\end{figure}

\begin{figure}[htbp]
    \centering
    \includegraphics[width=0.9\textwidth]{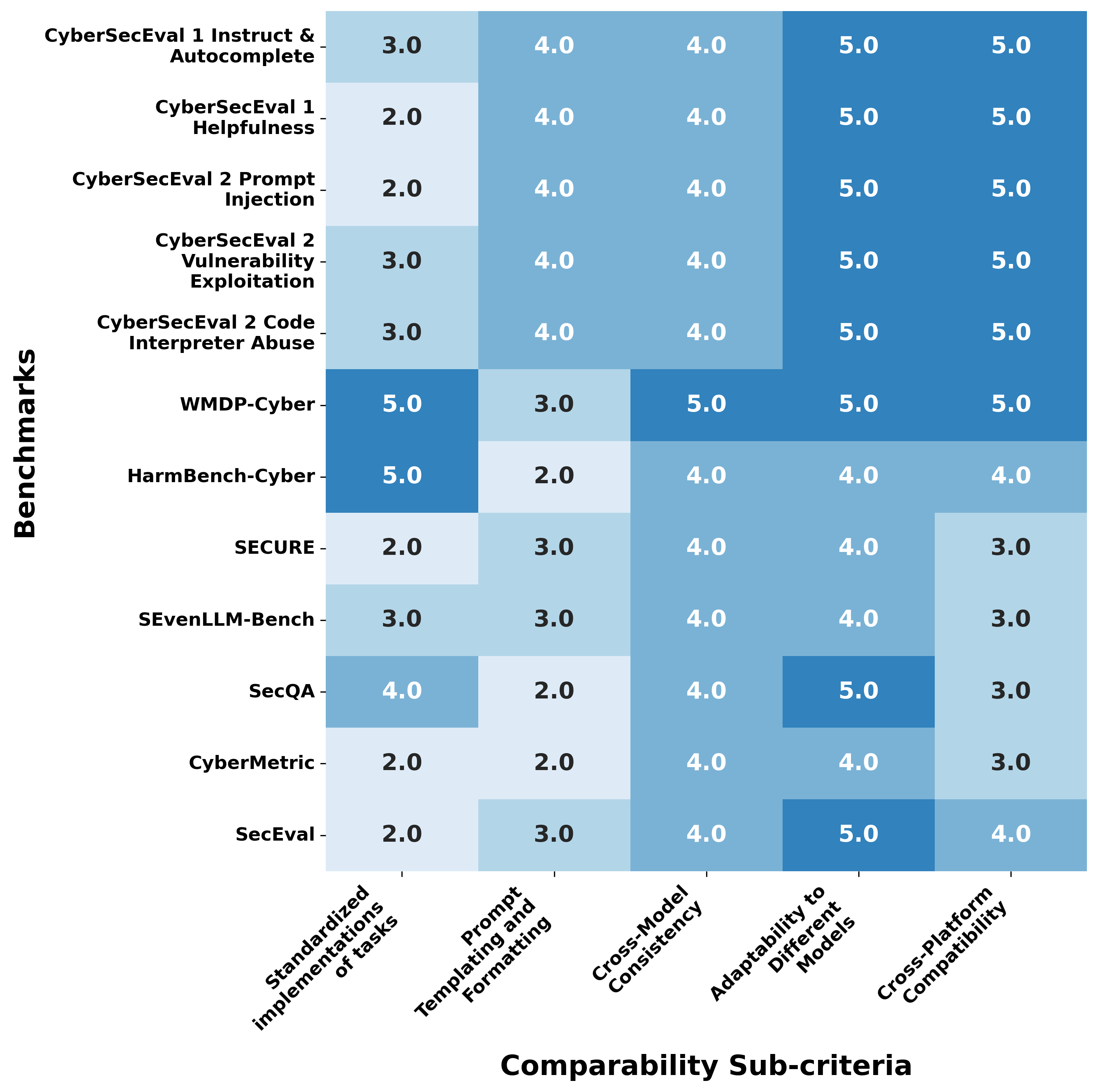}
    \caption{Scores of cybersecurity benchmarks across comparability sub-criteria.}
    \label{fig:comparability}
\end{figure}

\begin{figure}[htbp]
    \centering
    \includegraphics[width=0.9\textwidth]{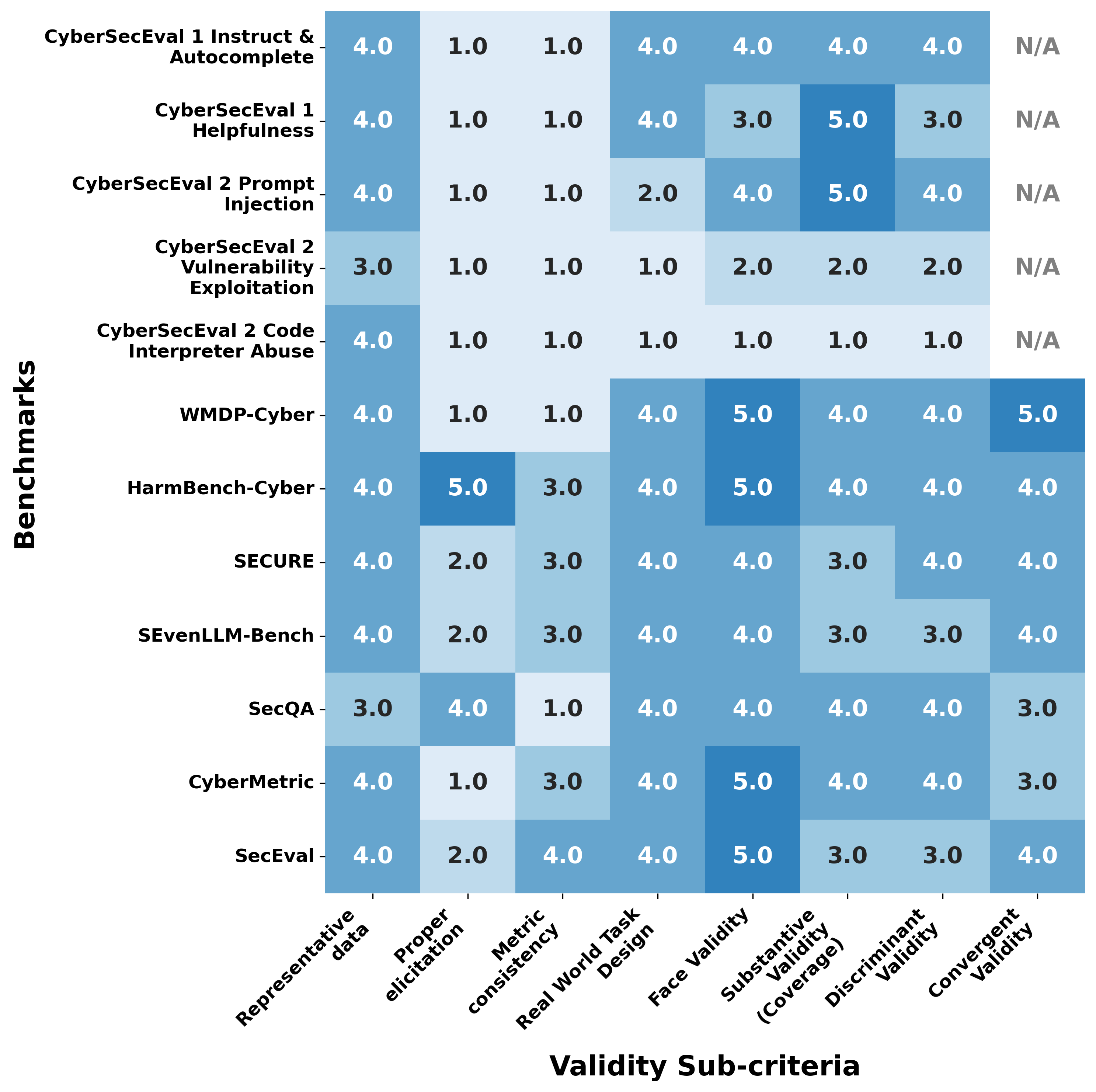}
    \caption{Scores of cybersecurity benchmarks across validity sub-criteria.}
    \label{fig:validity}
\end{figure}

\begin{figure}[htbp]
    \centering
    \includegraphics[width=0.9\textwidth]{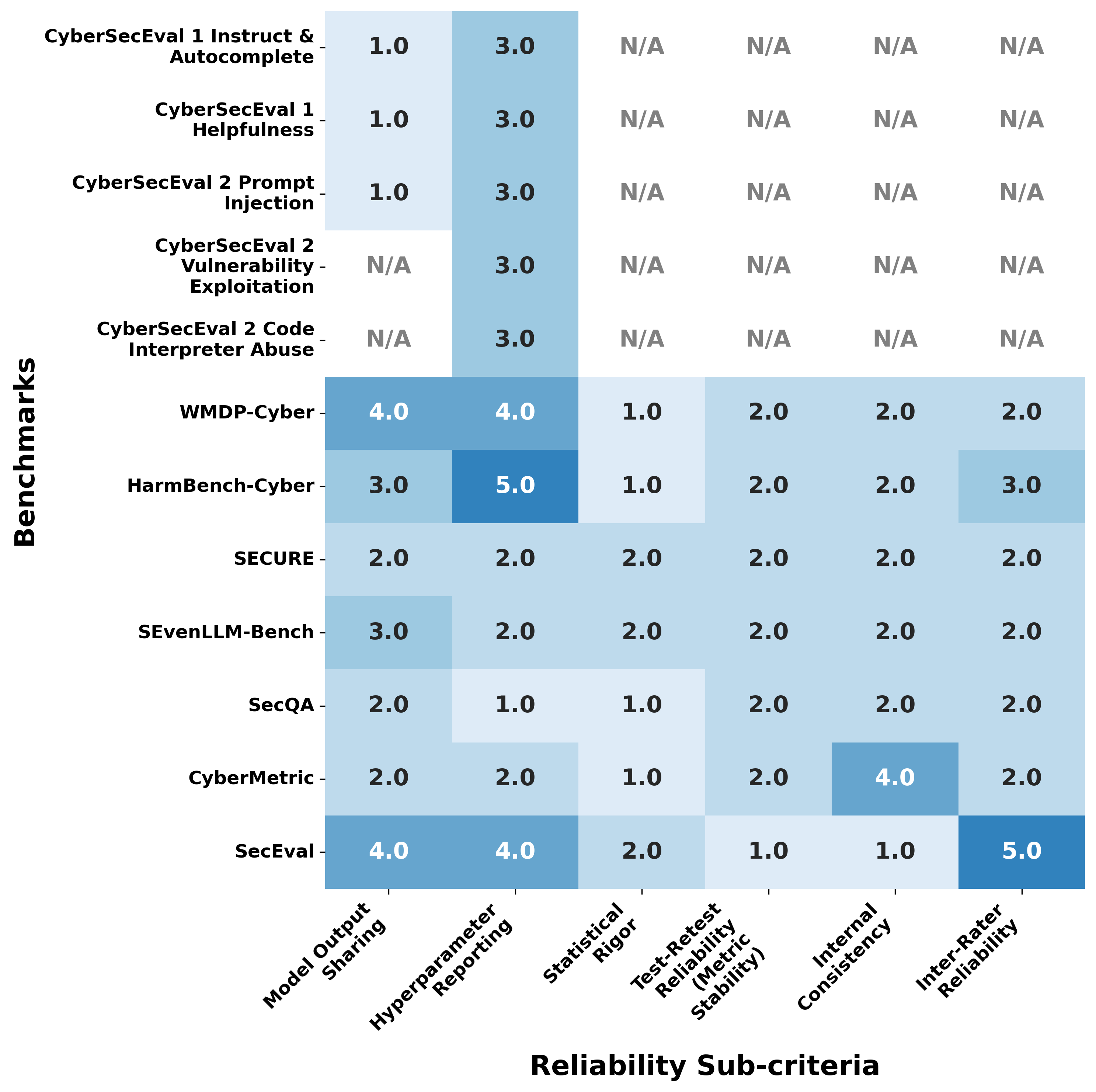}
    \caption{Scores of cybersecurity benchmarks across reliability sub-criteria.}
    \label{fig:reliability}
\end{figure}

In Figures~\ref{fig:memorization} to~\ref{fig:reliability}, we include scores from each of the sub-criteria. These scores are aggregated to create Figure~\ref{fig:scores}, Table~\ref{cybersec-eval-results} and Table~\ref{cybersec-eval-results1}.

\vfill
\pagebreak

\section{LLM evaluators}
As outlined in Section~\ref{section-2}, we tested MEQA using GPT-4o as an LLM evaluator. We attempted both zero-shot prompting and few-shot prompting as preliminary tests. In the latter case, we included scores for benchmarks that had been already evaluated by humans, this improved the LLM's accuracy. Our results in this paper are those obtained using few-shot prompting.

\section{Limitations}
\label{limitations}
There are several limitations to our work that we hope to address in the future. Fundamentally, our work is limited to QA benchmarks, we hope to expand this scope eventually. Furthermore, our tests were proof-of-concepts limited to three human evaluators and one LLM evaluators; we would like to test our framework with more human and LLM evaluators.

Some limitations are specific to the user of LLM evaluators. We found that GPT-4o sometimes struggles to provide correct, accurate scores. LLMs often learn biases from authors’ tone, inflating scores when the tone is positive. The automated scoring also fails in rare cases where the sub-criterion is not applicable. These issues may be fixed with better prompting; for example, for the case where the LLM fails to recognise the sub-criterion is inapplicable, we could attempt placing more emphasis in the prompt to check for applicability first, and then generate scores in a second prompt.

The prompt template we used is as follows:

\begin{itshape}
For a given benchmark we have to evaluate this criteria: [placeholder for sub-criteria]

Here is the explanation about how well the benchmark fits this criteria: [placeholder for explanation]

Based on this information, come up with a score from 1 to 5 (where 5 is the best fit) or N/A if the criteria does not apply to the benchmark. You should also provide a comment explaining your score and for that inspire yourself in the following examples. In the examples the same criteria is evaluated.

Here are a few examples to help you understand how to score this criteria:
[placeholder for few-shot examples]

Provide your response in the following format:
\newline Score: [Your score]
\newline Final Comment: [Your comment explaining the score]
\end{itshape}

\pagebreak 

\section{Additional scores}
Table~\ref{cybersec-eval-results} lists mean scores with standard deviations for each benchmark, our initial analysis broke down CyberSecEval 1 and 2 into their constituent sections to understand if performance varied between them. We include the full list of mean scores and standard deviations in Table~\ref{cybersec-eval-results1}.

\label{error-analysis}
\begin{table}[H]
  \caption{Cybersecurity Evaluation Results}
  \label{cybersec-eval-results1}
  \centering
  \begin{tabular}{@{}lr@{}}
    \toprule
    \textbf{Evaluation} & \textbf{Score}\\
    \midrule
    CyberSecEval 1 Instruct \& Autocomplete & 3.2 \(\pm\) 1.4 \\
    \addlinespace[0.5ex]
    CyberSecEval 1 Helpfulness & 3.2 \(\pm\) 1.5 \\
    \addlinespace[0.5ex]
    CyberSecEval 2 Prompt Injection & 3.6 \(\pm\) 1.4 \\
    \addlinespace[0.5ex]
    CyberSecEval 2 Vulnerability Exploitation & 3.3 \(\pm\) 1.5 \\
    \addlinespace[0.5ex]
    CyberSecEval 2 Code Interpreter Abuse & 3.2 \(\pm\) 1.7 \\
    \addlinespace[0.5ex]
    WMDP-Cyber & 3.5 \(\pm\) 1.6 \\
    \addlinespace[0.5ex]
    HarmBench-Cyber & 3.6 \(\pm\) 1.3 \\
    \addlinespace[0.5ex]
    SECURE & 2.7 \(\pm\) 1.1 \\
    \addlinespace[0.5ex]
    SEvenLLM-Bench & 2.9 \(\pm\) 1.0 \\
    \addlinespace[0.5ex]
    SecQA & 2.7 \(\pm\) 1.2 \\
    \addlinespace[0.5ex]
    CyberMetric & 2.8 \(\pm\) 1.2 \\
    \addlinespace[0.5ex]
    SecEval & 3.2 \(\pm\) 1.4 \\
    \bottomrule
  \end{tabular}
\end{table}

\section{Experimental details and resources}
\label{experimental-details}
Our experiments using GPT-4o as an LLM evaluator were run on a HP Probook (32GB RAM, i5). Our code took approximately 10 minutes to run.

\end{document}